\documentclass{article}
\PassOptionsToPackage{table}{xcolor}

\usepackage{microtype}
\usepackage{graphicx}
\usepackage{subcaption}
\usepackage{booktabs} 

\usepackage{hyperref}

\usepackage[preprint]{icml2026}

\usepackage{amsmath}
\usepackage{amssymb}
\usepackage{mathtools}
\usepackage{amsthm}

\usepackage[capitalize,noabbrev]{cleveref}

\usepackage{multirow}
\usepackage{booktabs}
\usepackage{tabularx}
\newcolumntype{C}{>{\centering\arraybackslash}X}
\newcommand{\myhl}[1]{\cellcolor{red!10}#1}
\newcommand{\myhlb}[1]{\cellcolor{red!10}\textbf{#1}}

\theoremstyle{plain}

\theoremstyle{definition}

\theoremstyle{remark}

\usepackage[textsize=tiny]{todonotes}

\icmltitlerunning{SONIC: Segmented Optimized Nexus for Information Compression in Key-Value Caching}

\begin{document}

\twocolumn[
  \icmltitle{SONIC: Segmented Optimized Nexus for\\
     Information Compression in Key-Value Caching}



  \icmlsetsymbol{equal}{*}

  \begin{icmlauthorlist}
    \icmlauthor{Hong Chen}{equal,hkustgz}
    \icmlauthor{Xiang Liu}{equal,hkustgz}
    \icmlauthor{Bo Wang}{hkustgz}
    \icmlauthor{Yuxuan Fan}{hkustgz}
    \icmlauthor{Yuanlin Chu}{hkustgz}
    \icmlauthor{Zongluo Li}{kcl}
    \icmlauthor{Xiaowen Chu}{hkustgz}
    \icmlauthor{Xuming Hu}{hkustgz}
  \end{icmlauthorlist}

  \icmlaffiliation{hkustgz}{The Hong Kong University of Science and Technology (Guangzhou), China}
  \icmlaffiliation{kcl}{King's College London, United Kingdom}

  \icmlcorrespondingauthor{Xiaowen Chu}{xwchu@hkust-gz.edu.cn}
  \icmlcorrespondingauthor{Xuming Hu}{xuminghu@hkust-gz.edu.cn}

  \icmlkeywords{Machine Learning, ICML}

  \vskip 0.3in
]



\printAffiliationsAndNotice{\icmlEqualContribution}

\begin{abstract}
  The linear growth of Key-Value (KV) cache remains a bottleneck for multi-turn LLM deployment. Existing KV cache compression methods often fail to account for the structural properties of multi-turn dialogues, relying on heuristic eviction that risks losing critical context. We propose \textbf{SONIC}, a learning-based framework that compresses historical segments into compact and semantically rich \textbf{Nexus} tokens. By integrating dynamic budget training, SONIC allows flexible adaptation to varying memory constraints without retraining. Experiments show that at compression ratios of 80\% and 50\%, SONIC consistently outperforms baselines such as H2O and StreamingLLM on four diverse multi-turn benchmarks. Specifically, on the widely used MTBench101 benchmark, SONIC achieves an average score improvement of 35.55\% over state-of-the-art baselines, validating its effectiveness in sustaining coherent multi-turn dialogues. Furthermore, SONIC enhances deployment efficiency, accelerating the overall inference process by 50.1\% compared to full-context generation.
\end{abstract}

\section{Introduction}
Multi-turn conversation has emerged as a dominant application for Large Language Models (LLMs). However, the standard self-attention mechanism in Transformers requires maintaining a monotonically growing Key-Value (KV) cache~\cite{vaswani2017attention}. Consequently, inference memory and latency scale linearly with context length. Although approaches like FlashAttention~\cite{dao2022flashattention} have optimized memory access patterns (IO-aware), the linear growth of the KV cache remains a critical bottleneck for efficient multi-turn generation.

To address the long-context challenge, researchers have proposed architectural alternatives such as linear attention and State Space Models~\cite{wang2020linformer,choromanski2020rethinking, gu2024mamba}. However, the Transformer remains the dominant architecture for mainstream dialogue models. High retraining costs and limited ecosystem compatibility have restricted the large-scale deployment of these alternatives. Conversely, while inference-time KV cache compression strategies like H2O~\cite{zhang2023h2o}, SnapKV~\cite{li2024snapkv}, and StreamingLLM~\cite{xiao2023efficient} effectively reduce memory usage, they were primarily designed for single-turn long-context scenarios. Consequently, these methods often sacrifice cross-turn dependencies and fail to meet the dynamic adaptation requirements of multi-turn interactions. Recent evaluations indicate that such aggressive compression significantly degrades model performance on standard benchmarks~\cite{liu2025can}. Furthermore, in multi-turn settings, these approaches suffer from progressive information loss due to the repetitive compression of historical context~\cite{liu2025flowkv}.

\begin{figure}[t]
    \centering
    \includegraphics[width=0.9\linewidth]{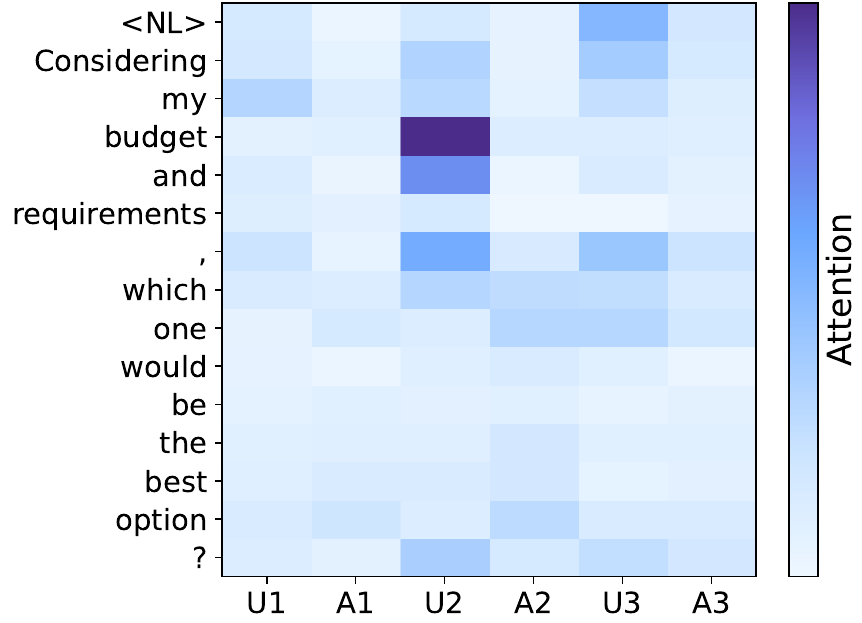}
    \caption{Attention Pattern Visualization of SONIC model on MTBench101 Sample. U$X$ and A$X$ indicate the $X$-th turn of user and assistant messages, respectively. The original Tokens of historical rounds have been discarded and compressed into Nexus Tokens.}
    \label{fig:attention}
\end{figure}

Therefore, we propose \textbf{SONIC} (\textbf{\underline{S}}egmented \textbf{\underline{O}}ptimized \textbf{\underline{N}}exus for \textbf{\underline{I}}nformation \textbf{\underline{C}}ompression), a novel framework for KV cache eviction. At the heart of SONIC lies the Nexus mechanism, which uses specialized Nexus tokens to aggregate and compress historical context. By leveraging turn-level and position-level representations, the model distinguishes between different historical segments. Furthermore, we combine hierarchical visibility constraints with KV compression rules to ensure that only the system prompt, Nexus tokens, and the current query are retained. This approach effectively maps long-term history into a compact set of retrievable representations, striking a balance between inference efficiency and generation performance. As an example shown in~\autoref{fig:attention}, we observe that even when raw historical text is discarded, the model explicitly attends to the Nexus tokens of the past turn (e.g., U2) where the budget details were originally located to retrieve specific constraints during current generation, validating the semantic integrity of our compression.

Furthermore, we introduce a dynamic budget training strategy. By randomly sampling the Nexus budget during the training phase, the model acquires the ability to adaptively adjust compression ratios based on available memory at inference time, eliminating the need for retraining. Beyond standard multi-turn datasets such as SafeDialBench~\cite{cao2025safedialbenchfinegrainedsafetybenchmark} and MTBench101~\cite{bai2024mt}, we construct two specialized benchmarks to evaluate long-term memory and topic shifting capabilities: GSM8K-Variant (for multi-turn topic-shifting mathematical tasks) and CoreRes (for multi-turn coreference resolution).

Our contributions are summarized as follows: 
\begin{itemize} 
    \item \textbf{Structured Memory Mechanism:} We propose SONIC, a learnable compression method equipped with hierarchical visibility constraints on Nexus tokens. This mechanism achieves efficient, learnable context compression tailored for multi-turn conversations. 
    \item \textbf{Integrated Training Scheme:} We design a dual-objective training framework combining distillation and reconstruction. This ensures that the compressed representations maintain sufficient semantic density to preserve critical context. 
    \item \textbf{Dynamic Budget Training:} We introduce a dynamic budget strategy that randomizes Nexus allocation during training, endowing the model with robust inference capabilities across varying memory constraints. 
    \item \textbf{Systematic Benchmark Evaluation:} We construct two specialized benchmarks (GSM8K-Variant and CoreRes) and conduct a comprehensive evaluation across MTBench101 and SafeDialBench tasks, demonstrating the effectiveness of our method in complex multi-turn scenarios. 
\end{itemize}

\section{Related Works}
The linear memory complexity of the Transformer's Key-Value (KV) cache poses a significant bottleneck for long-context generation. To mitigate this, recent research has diverged into two main streams: architectural modifications and inference-time compression. Architectural approaches, such as Linear Attention~\cite{wang2020linformer,choromanski2020rethinking} and State Space Models like Mamba~\cite{gu2024mamba}, achieve sub-quadratic complexity but often require expensive retraining and suffer from ecosystem fragmentation. In contrast, inference-time compression strategies operate on standard Transformers. Heuristic eviction methods, such as H2O~\cite{zhang2023h2o}, discard tokens based on accumulated attention scores. StreamingLLM~\cite{xiao2023efficient} identifies the importance of initial tokens (``attention sinks'') and maintains a sliding window of recent tokens.

\section{SONIC Framework}
\begin{figure*}
    \centering
    \includegraphics[width=0.9\linewidth]{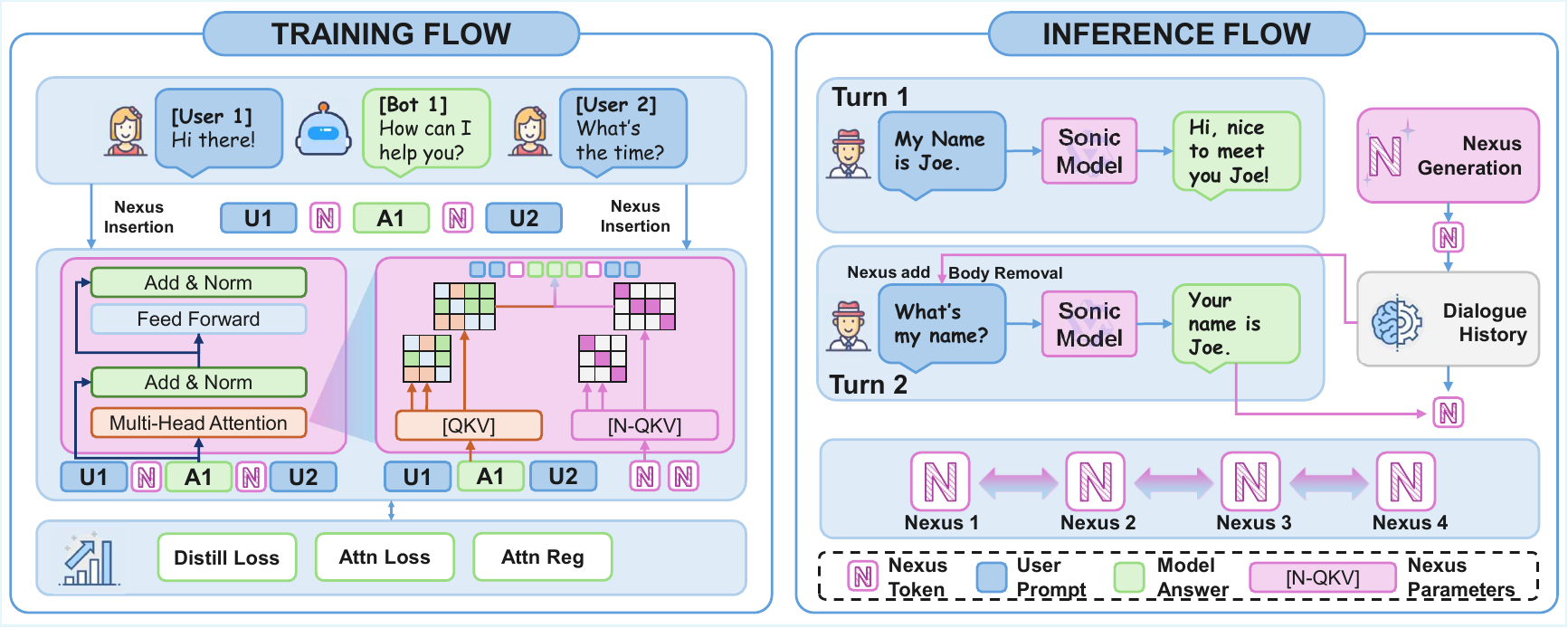}
    \caption{Overview of the SONIC framework. Historical segments are compressed into Nexus tokens, which are fully connected across turns. The hierarchical visibility mask ensures that only the system prompt, Nexus tokens, and the current query are accessible during generation.}
    \label{fig:sonic_overview}
\end{figure*}

\textbf{SONIC} (\textbf{\underline{S}}egmented \textbf{\underline{O}}ptimized \textbf{\underline{N}}exus for \textbf{\underline{I}}nformation \textbf{\underline{C}}ompression) is a comprehensive framework designed to resolve the memory bottleneck in multi-turn LLM interactions, as illustrated in \cref{fig:sonic_overview}. The core of this framework is the \textbf{Nexus}, a set of learnable tokens inserted into historical segments to act as information aggregates. While SONIC defines the overall methodology for compression and retrieval, Nexus serves as the tangible medium for storing and accessing these compressed memories, enabling the model to retain long-term context with minimal memory footprint.

\subsection{Nexus Insertion and Representation}
We decompose the multi-turn conversation sequence into discrete message segments, delimited by control tokens (e.g., \texttt{<|im\_start|>}, \texttt{<|im\_end|>}) and role identifiers to distinguish between the system, user, and assistant. We define the most recent user query as the \textit{current query}, while all preceding non-system segments constitute the \textit{historical context}. Crucially, Nexus insertion is applied exclusively to this historical context. This ensures that only historical information undergoes compression, while the system prompt and the current query are preserved in their original, uncompressed state.

Formally, we treat each user utterance and assistant response as an individual historical segment $m_i$. For each segment, a set of $K$ Nexus tokens is appended. The augmented input sequence is constructed as follows:
\begin{equation}
    \mathcal{S}_{\text{aug}} = \{ I_{\text{sys}}, m_1, \mathcal{N}^{(1)}, m_2, \mathcal{N}^{(2)}, \ldots, m_h, \mathcal{N}^{(h)}, Q_{\text{curr}} \},
\end{equation}
where $h$ denotes the number of historical segments, $Q_{\text{curr}}$ is the current query, and $\mathcal{N}^{(i)} = [N_1^i, \ldots, N_K^i]$ represents the sequence of Nexus tokens inserted after the user or assistant messages of the $i$-th segment. We treat a pair of user and assistant messages as a single conversation turn, assigning an identical turn ID to all Nexus tokens within the same turn.

Nexus tokens utilize a shared, learnable base embedding $\mathbf{e}_{\text{base}}$, initialized by averaging the embeddings of semantic compression keywords (e.g., ``summary'', ``condensed''), rather than expanding the tokenizer vocabulary. The embedding of the $j$-th Nexus token at turn $t$ is composed of three components:
\begin{equation}
    \mathbf{e}(N_{j}^{t}) = \mathbf{e}_{\text{base}} + \mathbf{e}_{\text{pos}}[j] + \mathbf{e}_{\text{turn}}[t],
\end{equation}
where $\mathbf{e}_{\text{pos}}[j]$ ($j \in \{1, \ldots, K\}$) and $\mathbf{e}_{\text{turn}}[t]$ denote the position embedding and turn embedding, respectively. The position embedding enables distinct Nexus tokens to capture different semantic aspects within a segment, while the turn embedding allows the model to differentiate compressed information across distinct temporal contexts.

To further specialize Nexus tokens for compression tasks, we decouple their parameters from standard text tokens. This is achieved by employing dedicated projection matrices (e.g., $W_Q^N, W_K^N, W_V^N, W_O^N$) and independent MLP branches within the Transformer blocks. This design ensures that Nexus tokens focus on information aggregation (``compression'') rather than surface-level text generation. Finally, the parameter $K$ serves as a flexible budget constraint; the subsequent dynamic budget training strategy aims to enhance model robustness across varying configurations of $K$.

\subsection{Hierarchical Visibility Mask}
To ensure consistency between training and inference regarding context visibility, we construct a hierarchical attention mask on the augmented sequence. Let $L$ denote the length of the augmented sequence. We define an additive attention mask $M \in \mathbb{R}^{L \times L}$, where $M_{p,q} = 0$ indicates that token $q$ is visible to token $p$, and $M_{p,q} = -\infty$ otherwise.

For each historical segment $m_i$, we define two sets of position indices:
\begin{itemize}
    \item $\mathcal{B}_i$: The set of indices corresponding to the original text tokens (body) of the $i$-th segment.
    \item $\mathcal{N}_i = \{N_1^i, \ldots, N_K^i\}$: The set of indices corresponding to the Nexus tokens inserted at the $i$-th segment.
\end{itemize}

Based on these definitions, the mask $M$ is constructed according to the following three rules:

\noindent\textbf{Basic Causal Rule.} Standard causal masking is applied globally to ensure that no token can attend to future text tokens, preserving the autoregressive property. For any pair of non-Nexus tokens at positions $p$ and $q$:
\begin{equation}
    M_{p,q} = 0 \quad \text{if and only if} \quad q \le p,
\end{equation}
otherwise $M_{p,q} = -\infty$.

\noindent\textbf{Historical Body Removal.} To enforce compression, the raw text of a historical segment becomes inaccessible once its corresponding Nexus tokens are generated. Specifically, for any historical segment $m_i$, if the current query position $p$ is located after the last Nexus token of this segment (i.e., $p > \max(\mathcal{N}_i)$), the body tokens $\mathcal{B}_i$ are masked:
\begin{equation}
    \forall q \in \mathcal{B}_i, \quad M_{p,q} = -\infty.
\end{equation}
Consequently, in subsequent turns (including the current generation phase), the model can only access the system prompt and the Nexus anchors of past segments. The compressed message body cannot be ``peeked at'', thereby compelling the model to rely solely on the Nexus representations for historical context.

\noindent\textbf{Bidirectional Nexus Visibility.} We eliminate causal constraints among Nexus tokens to facilitate global information flow. Within the same segment, Nexus tokens are bidirectionally visible: $\forall a, b \in \mathcal{N}_i, M_{a,b} = 0$. Furthermore, across different historical segments, Nexus tokens are fully visible to each other:
\begin{equation}
    \forall a \in \mathcal{N}_i, b \in \mathcal{N}_j, \quad M_{a,b} = 0.
\end{equation}
This mechanism allows Nexus tokens to form a fully connected memory graph across the entire conversation history, while standard text tokens continue to adhere to the local causal structure.

\section{Training Strategy}
\subsection{Multi-Level Knowledge Distillation}
Since the compression of context into the Nexus state $N$ inherently incurs information loss, we employ a teacher-student knowledge distillation framework. The teacher model processes the full, uncompressed context, whereas the student model operates on the sequence with Nexus compression applied. To ensure structural alignment, distillation losses are computed exclusively at positions where a one-to-one correspondence exists between the teacher and student sequences (i.e., the set of indices $\Omega$ corresponding to the system prompt and the current turn).

\noindent\textbf{Logit Distillation.} Let $p_T(\cdot \mid x)$ and $p_S(\cdot \mid x)$ denote the output probability distributions of the teacher and student models at an aligned position $x \in \Omega$, respectively. With a temperature parameter $\tau$, the logit distillation loss is defined as:
\begin{equation}
    \mathcal{L}_{\text{KD}} = \tau^2 \sum_{x \in \Omega} \mathrm{KL}\left( p_T(\cdot \mid x; \tau) \;\|\; p_S(\cdot \mid x; \tau) \right).
\end{equation}

\noindent\textbf{Hidden-state Alignment.} To enforce deeper semantic alignment, we impose a cosine similarity constraint on the hidden states of a specified layer (typically the final layer). Let $h_T(x)$ and $h_S(x)$ represent the hidden states at position $x$:
\begin{equation}
    \mathcal{L}_{\text{H}} = \frac{1}{|\Omega|} \sum_{x \in \Omega} \left( 1 - \cos(h_T(x), h_S(x)) \right).
\end{equation}

\noindent\textbf{Attention-guided Distillation.} Mere distribution alignment is often insufficient for ensuring that Nexus anchors effectively ``remember'' critical history. We therefore utilize the teacher's attention patterns as importance priors. Specifically, we average the teacher's attention matrix across multiple heads to quantify the saliency of historical segments. We compute an importance score $w(x)$ for history tokens based on the accumulated attention weights they receive from the current response. This weight is then applied to the distillation objective:
\begin{equation}
    \mathcal{L}_{\text{AKD}} = \tau^2 \sum_{x \in \Omega} w(x) \cdot \mathrm{KL}\left( p_T(\cdot \mid x; \tau) \;\|\; p_S(\cdot \mid x; \tau) \right).
\end{equation}

\noindent\textbf{Nexus Attention Regularization.} To ensure that the model actively utilizes the compressed memory, we introduce a regularization term that penalizes low attention weights allocated to Nexus tokens. Let $A_{\text{nexus}}$ denote the average attention weight assigned to Nexus tokens by the current query. We define a hinge loss with a threshold $\gamma$:
\begin{equation}
    \mathcal{L}_{\text{Reg}} = \max(0, \gamma - A_{\text{nexus}}).
\end{equation}

\subsection{Information Bottleneck and Reconstruction}
The Nexus mechanism inherently functions as an information bottleneck: for each historical segment, the raw token body $\mathcal{B}_i$ is discarded, leaving only the Nexus tokens $\mathcal{N}_i$ accessible for future retrieval. To endow these compressed representations with retrievability, we introduce a reconstruction loss that compels each Nexus token to semantically encode its corresponding historical chunk.

Specifically, we partition the discarded body sequence $\mathcal{B}_i = \{b_1, \ldots, b_{|\mathcal{B}_i|}\}$ of segment $m_i$ into $K$ sub-intervals $\mathcal{B}_i^1, \ldots, \mathcal{B}_i^K$, corresponding one-to-one with the Nexus tokens $\mathcal{N}_i$. Let $h(\cdot)$ denote the model's hidden state. We define the reconstruction target $\tilde{h}_i^j$ for the $j$-th Nexus token as an attention-weighted aggregation of its corresponding sub-interval:
\begin{equation}
    \begin{aligned}
        \tilde{h}_i^j &= \sum_{x \in \mathcal{B}_i^j} \alpha(x) \, h(x), \\
        \alpha(x) &= \frac{\exp(\cos(h(N_j^i), h(x)))}{\sum_{y \in \mathcal{B}_i^j} \exp(\cos(h(N_j^i), h(y)))},
    \end{aligned}
\end{equation}
in cases where the segment length is shorter than $K$, we employ padding or dynamically reduce $K$ for that specific segment to avoid empty intervals.

The reconstruction loss minimizes the cosine distance between the Nexus embedding and this target representation. Furthermore, to prevent attention collapse, we incorporate an entropy regularization term:
\begin{equation}
    \begin{aligned}
    \mathcal{L}_{\text{recon}} = \frac{1}{|\mathcal{N}_{\text{total}}|} \sum_{i,j} \bigg[ \Big(1 &- \cos\big(h(N_j^i), \tilde{h}_i^j\big)\Big) \\
    &+ \beta \cdot \left(1 - \frac{H(\alpha)}{\log |\mathcal{B}_i^j|}\right) \bigg],
    \end{aligned}
\end{equation}
where $H(\alpha)$ represents the entropy of the attention weights, and $|\mathcal{N}_{\text{total}}|$ is the total number of Nexus tokens. This objective encourages Nexus tokens to capture the semantic content of their respective intervals broadly and accurately, serving as effective carriers for compressed memory.

\noindent\textbf{Total Loss.} Hence, the final training objective is a weighted sum of the above mentioned components:
\begin{equation}
    \begin{aligned}
    \mathcal{L}_{\text{total}} = \lambda_{\text{KD}} \mathcal{L}_{\text{KD}} + \lambda_{\text{H}} \mathcal{L}_{\text{H}} + \lambda_{\text{AKD}} \mathcal{L}_{\text{AKD}} & \\
    + \lambda_{\text{Reg}} \mathcal{L}_{\text{Reg}} + \lambda_{\text{Recon}} \mathcal{L}_{\text{Recon}}.
    \end{aligned}
\end{equation}

\subsection{Adaptive Budget Training}
To enable dynamic budget control during the inference phase, we explicitly incorporate a variable Nexus allocation mechanism into both the model architecture and the training pipeline. We define a maximum capacity $K_{\max}$ and a default quantity $K_0$ for the Nexus tokens per segment.

To ensure parameter consistency, the Nexus position embedding matrix is initialized with a fixed size of $K_{\max}$. When an active budget $K < K_{\max}$ is employed, we utilize only the first $K$ position embeddings. This approach ensures that representations under varying budgets are subsets of a unified parameter space, facilitating smooth transitions between different compression levels.

During training, we employ an Adaptive Budget Sampling strategy. For each training batch, we sample a budget $K$ from a pre-defined discrete set of candidates $\mathcal{K}_{set}$ (e.g., powers of 2) to encourage structured compression levels:
\begin{equation}
    K \sim \text{Uniform}(\mathcal{K}_{set}), \quad \text{where } \mathcal{K}_{set} \subseteq \{1, ..., K_{max}\}.
\end{equation}
This sampled $K$ is applied consistently across all historical segments within the batch, governing the entire pipeline—from Nexus insertion and visibility mask construction to KV compression and distillation alignment. Specifically, the parsing module propagates the sampled $K$ to all downstream components, ensuring that the reconstruction targets and loss calculations are strictly aligned with the current budget.

Formally, the training objective is to minimize the ExpectedAttention loss over both the data distribution $\mathcal{D}$ and the budget distribution:
\begin{equation}
    \min_{\theta} \; \mathbb{E}_{S \sim \mathcal{D}} \; \mathbb{E}_{K \sim \mathcal{U}(\mathcal{K}_{set})} \big[ \mathcal{L}(S; K) \big],
\end{equation}
where $\mathcal{L}(S; K)$ represents the weighted sum of the distillation and reconstruction losses defined in previous sections.

During inference, this mechanism provides flexibility: users can explicitly specify a budget $K$ to strictly satisfy memory or latency constraints. In the absence of a specific requirement, the model gracefully reverts to the default setting $K_0$.
\section{Benchmarks}
We evaluate our model across four distinct multi-turn tasks. These include two synthetic benchmarks designed to rigorously test long-term memory and topic-switching capabilities (see~\autoref{appendix:data} for construction details), as well as two established public benchmarks to assess general multi-turn proficiency. Collectively, these tasks cover core capabilities, including cross-turn reasoning, coreference resolution, multi-turn conversation safety, and overall dialogue quality.

\noindent\textbf{GSM8K-Variant.}
A multi-turn mathematical reasoning task where problem conditions~\cite{cobbe2021gsm8k} are decomposed and scattered across a conversation interleaved with chitchat. It evaluates the model's ability to aggregate fragmented information under noise.

\noindent\textbf{Coreference Resolution.}
A synthetic task designed to test ultra-long entity tracking. It requires the model to recall specific attributes of an entity introduced in early turns after extensive distraction dialogues, triggered by an implicit reference.

\noindent\textbf{SafeDialBench~\cite{cao2025safedialbenchfinegrainedsafetybenchmark}.}
SafeDialBench is a multi-turn dialogue safety benchmark designed to evaluate LLMs under realistic and sophisticated jailbreak scenarios. Each dialogue is designed to probe whether a model can consistently maintain safety alignment as constraints are progressively relaxed and harmful objectives are implicitly introduced. 

\noindent\textbf{MTBench101~\cite{bai2024mt}.}
MTBench101 serves as a comprehensive multi-turn open-ended dialogue benchmark, covering diverse task categories such as reasoning, creative writing, and role-playing. Adhering to the official evaluation protocol, we employ an LLM judge to score the final turn response and report both the overall average and category-specific scores. This benchmark quantifies the overall dialogue quality and generalization capability of the model after compression.
\section{Experiments}
\subsection{Settings}
We evaluate SONIC on the Qwen3 series (0.6B, 1.7B, and 4B)~\cite{qwen3technicalreport}. For each size, we distill a frozen teacher model into a student model, fine-tuning only the Nexus-related parameters (embeddings and projectors) on our synthesized multi-turn dataset. To enable flexible inference, we employ Adaptive Budget Training, evenly sampling the Nexus budget $K$ during training iterations. We compare SONIC against Full-context inference and state-of-the-art KV Cache Compression methods, including H2O, SnapKV, ExpectedAttention, and StreamingLLM, implemented via kvpress~\cite{devoto2025expectedattention}. Evaluation employs Qwen-max as a judge for subjective benchmarks (MTBench101, SafeDialBench) and standard metrics for objective tasks. Detailed hyperparameters, training configurations, and hardware specifications are provided in~\autoref{appendix:details}.

\subsection{Main Results}

\begin{table*}[t]
\centering
\small
\caption{Performance comparison of SONIC against various KV cache compression baselines across Qwen3-0.6B, 1.7B, and 4B models. We report results on four benchmarks under 80\% and 50\% compression ratios, where the higher value indicates more information is compressed. Full-context performance is provided as the upper bound.}
\begin{tabular}{lcl ccccc} 
\toprule
\textbf{Model} & \textbf{Comp. Rat.} & \textbf{Algorithm} & \textbf{MTBench101} & \textbf{CoreRes (\%)} & \textbf{GSM8K-Var. (\%)} & \textbf{SafeDialBench} \\
\midrule
\multirow{11}{*}{Qwen3-0.6B} 
& -- & Full-Context & 6.00 & 57.36 & 2.05 & 6.66 \\
\cmidrule(lr){2-7}
& \multirow{5}{*}{80\%} 
& StreamingLLM        & 2.74 & 0.00 & 1.60 & 6.12 \\
& & ExpectedAttention & 3.62 & 6.03 & 1.20 & 6.51 \\
& & H2O                & 3.66 & 0.00 & 0.00 & 6.39 \\
& & SnapKV             & 3.21 & 0.72 & 1.20 & 6.44\\
& & \myhlb{SONIC} (Ours)      & \myhlb{3.93} & \myhlb{23.92} & \myhlb{1.40} & \myhlb{6.52} \\
\cmidrule(lr){2-7}
\cmidrule(lr){2-7}
& \multirow{5}{*}{50\%} 
& StreamingLLM        & 3.58 & 0.00 & 2.00 & 6.55 \\
& & ExpectedAttention & 3.92 & 19.33 & 2.09 & 6.56\\
& & H2O                & 3.92 & 11.25 & 2.00 & 6.44\\
& & SnapKV             & 3.93 & 19.22 & 1.30 & 6.52\\
& & \myhlb{SONIC} (Ours)    & \myhlb{5.61} & \myhlb{29.45} & \myhlb{2.50} & \myhlb{6.61} \\

\midrule

\multirow{11}{*}{Qwen3-1.7B} 
& -- & Full-Context & 8.11 & 87.53 & 9.10 & 8.10 \\
\cmidrule(lr){2-7}
& \multirow{5}{*}{80\%} 
& StreamingLLM        & 3.77 & 0.00 & 1.70 & 7.52 \\
& & ExpectedAttention & 4.74 & 6.03 & 2.40 & 7.60 \\
& & H2O                & 4.74 & 0.00 & 0.00 & 7.61 \\
& & SnapKV             & 4.21 & 0.72 & 1.70 & 7.62 \\
& & \myhlb{SONIC} (Ours)    & \myhlb{6.32} & \myhlb{24.07} & \myhlb{3.14} & \myhlb{7.66} \\
\cmidrule(lr){2-7}
& \multirow{5}{*}{50\%} 
& StreamingLLM        & 4.71 & 0.00 & 2.20 & 7.80 \\
& & ExpectedAttention & 4.84 & 45.09 & 4.50 & 7.85 \\
& & H2O                & 4.88 & 11.25 & 0.98 & 7.88 \\
& & SnapKV             & 4.76 & 19.22 & 4.10 &  7.90\\
& & \myhlb{SONIC} (Ours)    & \myhlb{6.42} & \myhlb{49.48} & \myhlb{7.00} & \myhlb{7.99} \\

\midrule

\multirow{11}{*}{Qwen3-4B} 
& -- & Full-Context & 9.03 & 94.79 & 35.44 & 8.42 \\
\cmidrule(lr){2-7}
& \multirow{5}{*}{80\%} 
& StreamingLLM        & 4.11 & 0.00 & 0.90 & 8.10 \\
& & ExpectedAttention & 5.02 & 9.20 & 4.50 & 8.18 \\
& & H2O                & 5.00 & 0.00 & 0.00 & 8.07\\
& & SnapKV             & 4.31 & 2.56 & 2.19 & 8.11\\
& & \myhlb{SONIC} (Ours)    & \myhlb{8.30} & \myhlb{68.40} & \myhlb{7.69} & \myhlb{8.27} \\
\cmidrule(lr){2-7}
& \multirow{5}{*}{50\%} 
& StreamingLLM        & 4.98 & 0.00 & 2.79 & 8.30\\
& & ExpectedAttention & 5.06 & 45.00 & 7.59 & 8.29\\
& & H2O                & 5.04 & 48.06 & 8.00 & 8.28\\
& & SnapKV             & 5.00 & 46.73 & 7.39 & 8.31\\
& & \myhlb{SONIC} (Ours)    & \myhlb{8.43} & \myhlb{69.01} & \myhlb{8.30} & \myhlb{8.33} \\
\bottomrule
\end{tabular}
\label{tab:main-results}
\end{table*}

We conducted experiments on four multi-turn conversation benchmarks: MTBench101~\cite{bai2024mt}, CoreRes, GSM8K-Variant~\cite{cobbe2021gsm8k}, and SafeDialBench~\cite{cao2025safedialbenchfinegrainedsafetybenchmark}. These benchmarks encompass long-context reasoning, coreference resolution, multi-turn conversation safety, and general conversation quality. Performance comparisons between the proposed SONIC method, the full-context baseline, and general KV cache compression methods are presented in~\autoref{tab:main-results}.

\textbf{MTBench101. }This benchmark utilizes an LLM-as-a-Judge mechanism to assign scores ranging from 0 to 10. The final metric is derived by calculating the minimum score across turns for each sample and then averaging these values across dimensions. SONIC consistently outperforms all KV cache compression baselines across both compression ratios and all model sizes. Specifically, at an 80\% compression ratio, SONIC achieved scores of 3.93 and 6.32 on Qwen3-0.6B and Qwen3-1.7B, respectively. These improvements are even more pronounced at a more moderate compression ratio of 50\%. Since MTBench101 evaluates performance based on the minimum score across turns, it is highly sensitive to single-turn errors. While general KV cache compression methods are prone to losing information from early turns, resulting in significant penalties, SONIC effectively retains cross-turn semantics through the learnable Nexus token. Notably, the performance gains scale positively with the increase in model size.

\textbf{CoreRes. }This task evaluates the model's capability in ultra-long-context entity tracing. Specifically, evaluation is based on whether the model can correctly identify entity-attribute relationships defined in the initial turns when queried at the final turn of the conversation. While most KV cache compression baselines degrade to near-zero accuracy, SONIC maintains robust performance, achieving scores of 0.2392 on Qwen3-0.6B and 0.6840 on Qwen3-4B. This performance gap stems from the limitations of general compression methods, such as StreamingLLM and H2O, which rely heavily on sliding windows or accumulated attention scores for token eviction. These strategies typically fail to account for the long-term utility of early-turn information, inadvertently discarding critical data necessary for resolving long-range dependencies.

\textbf{GSM8K-Variant. }In this benchmark, a single GSM8K problem is decomposed into multiple segments and presented across the initial turns of the dialogue. This is followed by a shift to unrelated topics before the final query solicits the solution. Consequently, the model must aggregate scattered conditions to derive the correct answer. Similar to CoreRes, general KV cache compression methods suffer severe performance degradation at an 80\% compression ratio. Although a performance gap remains between SONIC and the full-context upper bound, SONIC consistently outperforms the competing baselines. Notably, this task exhibits a more pronounced sensitivity to compression ratios and model sizes compared to other benchmarks.

\textbf{SafeDialBench. }This benchmark employs an LLM-as-a-judge mechanism to evaluate responses across three dimensions: Risk Identification, Risk Handling, and Consistency. The final metric is computed by aggregating the minimum score of each dimension across all conversation turns, followed by averaging these three values. This stringent scoring protocol implies that a safety failure in any single turn directly penalizes the overall score. Under an 80\% compression ratio with Qwen3-0.6B, SONIC achieves a score of 6.52, closely approaching the full-context baseline (6.66) and surpassing general KV cache compression baselines. General KV compression strategies may maintain safety standards in early turns. However, they often suffer from performance degradation in later stages due to context accumulation, leading to lower overall sample scores.

\begin{figure}[t]
    \centering
    \includegraphics[width=0.8\linewidth]{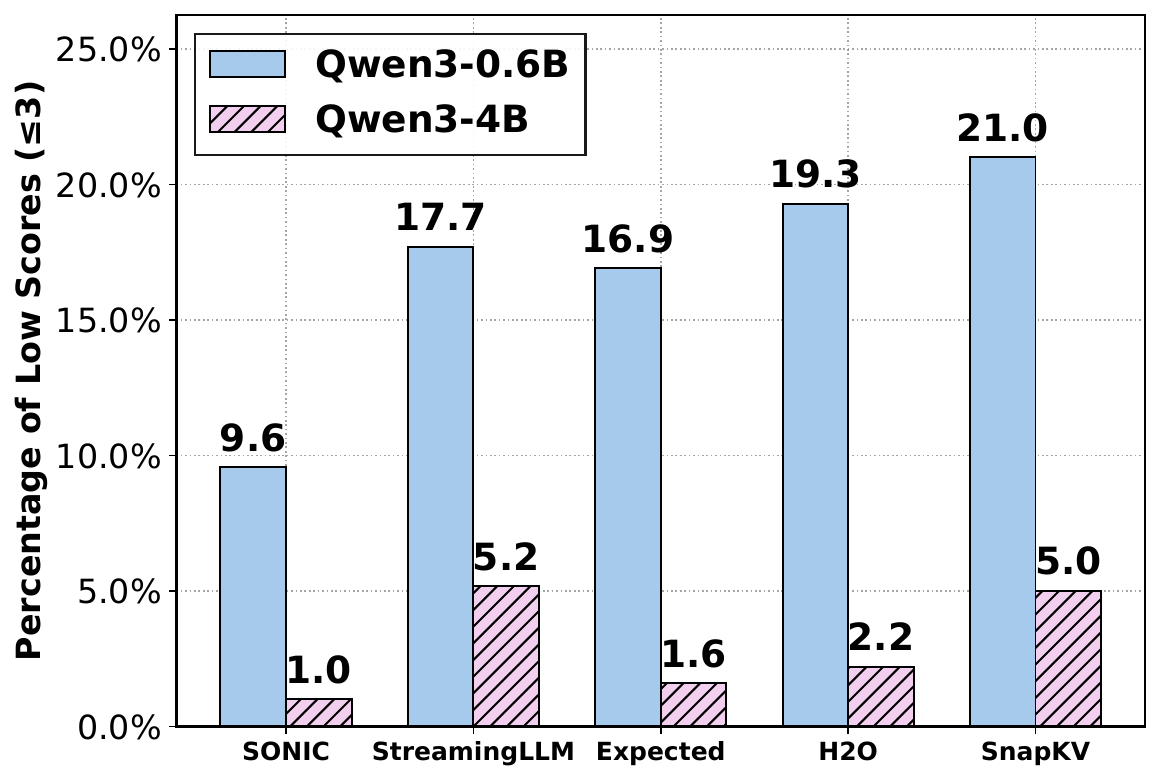}
    \caption{Comparison of Low-score Proportions (scores 1–3) for Turns $\ge$ 4 on MTBench101. Results are shown for Qwen3-0.6B and Qwen3-4B models at an 80\% compression rate. Lower proportions indicate superior multi-turn stability}
    \label{fig:mtbench-compare}
\end{figure}

In summary, the advantages of SONIC become most distinct as the compression ratio decreases, demonstrating its ability to approximate full-context fidelity under high-retention settings. While increasing model scale universally lifts performance, standard KV compression methods remain bottlenecked by their inability to preserve comprehensive information within complex multi-turn structures. Our statistical analysis, shown as~\autoref{fig:mtbench-compare} of MTBench-101 (focusing on late-stage interactions where turn $\geq 4$), reveals that standard KV compression strategies yield a significantly higher proportion of low scores ($\leq 3$) compared to SONIC as the conversation depth increases. Specifically, for the 0.6B model, SONIC maintains a low-score ratio of only 9.56\%, whereas StreamingLLM, ExpectedAttention, H2O, and SnapKV exhibit higher ratios of 17.69\%, 16.90\%, 19.28\%, and 21.00\%, respectively. A similar trend is observed in the 4B scale, where SONIC’s low-score ratio drops to 1.00\%, outperforming its counterparts (5.17\%, 1.59\%, 2.19\%, and 5.01\%). These results suggest that in extended multi-turn dialogues, conventional strategies—lacking explicit awareness of turn-based structures—are more prone to erroneous compression or loss of critical historical information. In contrast, SONIC maintains a significantly thinner ``low-score tail," demonstrating superior consistency and robustness across varying tasks and model scales.

\subsection{Ablation Study}
\subsubsection{Impact of Training Objectives}

\begin{table}[t]
\centering
\caption{Ablation Study of Training Objectives on CoreRes. We report the accuracy of the Qwen3-0.6B model at an 80\% compression ratio. The results highlight the contribution of each component: Nexus Attention Regularization ($\mathcal{L}_{\text{Reg}}$), Attention-Guided Distillation ($\mathcal{L}_{\text{AKD}}$), Hidden-state Alignment ($\mathcal{L}_{\text{H}}$), and Nexus Reconstruction ($\mathcal{L}_{\text{Recon}}$). }
\begin{tabular}{lcr}
\toprule
\textbf{Configuration} & \textbf{CoreRes (\%)} $\uparrow$ & \textbf{$\Delta$} (\%) \\
\midrule
\textbf{Full SONIC} & \textbf{23.92} & - \\
w/o Nexus-Attn-Reg & 13.17 & -10.75 \\
w/o Attn-Guide Distill & 14.52 & -9.40 \\
w/o Hidden-state Align & 15.44 & -8.48 \\
w/o Nexus-Recon & 21.66 & -2.26 \\
\bottomrule
\end{tabular}
\label{tab:ablation_loss}
\end{table}

We evaluate the contribution of each loss component ($\mathcal{L}_{\text{AKD}}$, $\mathcal{L}_{\text{H}}$, $\mathcal{L}_{\text{Reg}}$, $\mathcal{L}_{\text{Recon}}$) by removing them individually from the Qwen3-0.6B model at an 80\% compression ratio. As shown in~\autoref{tab:ablation_loss}, removing any component degrades performance, validating the necessity of the full objective. Notably, removing $\mathcal{L}_{\text{Reg}}$ yields the largest drop, indicating that explicitly enforcing attention to Nexus tokens is crucial to prevent the model from ignoring compressed context. Furthermore, the drops from removing $\mathcal{L}_{AKD}$ and $\mathcal{L}_{H}$ suggest that standard distillation alone is insufficient for deep semantic alignment. Finally, while the impact of removing $\mathcal{L}_{\text{Recon}}$ is smaller—likely due to implicit supervision from distillation—it remains essential for maintaining semantic integrity under high compression.

\subsubsection{Efficacy of Adaptive Budget Training}

\begin{figure}[t]
    \centering
    \includegraphics[width=0.9\linewidth]{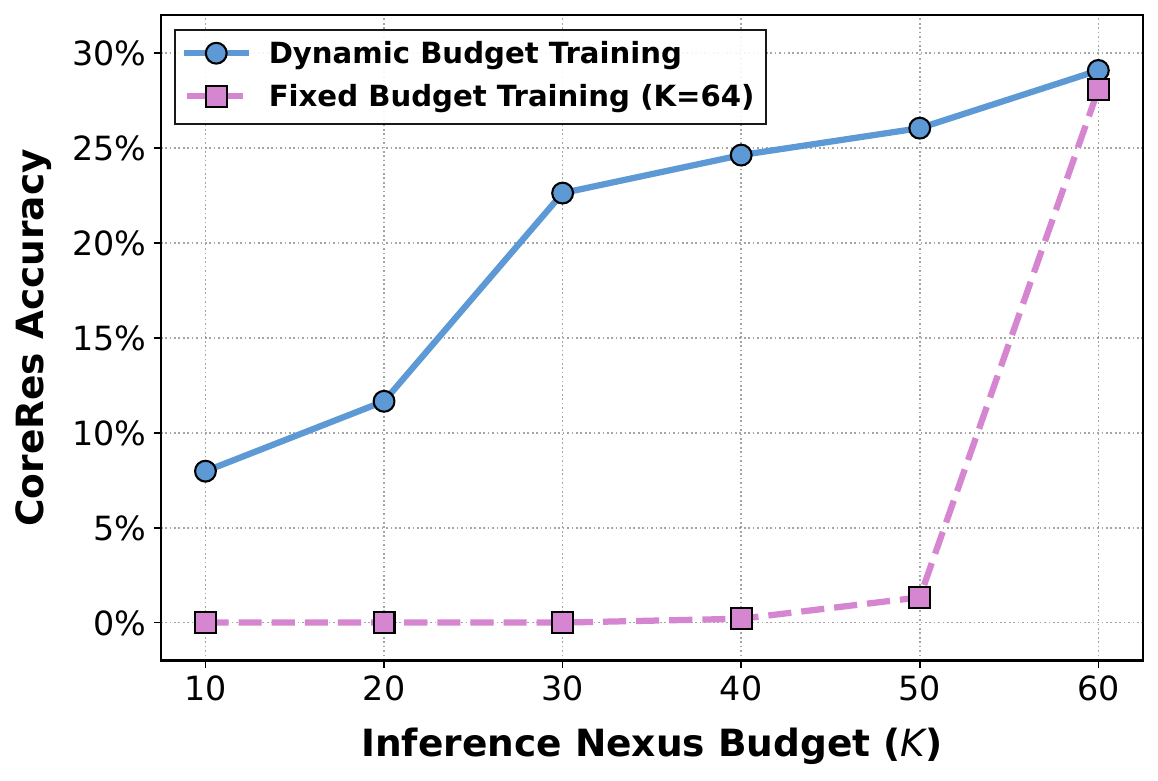}
    \caption{Ablation Study on Impact of Adaptive Budget Training on model robustness. We evaluate the Qwen3-0.6B model on the CoreRes benchmark under an 80\% compression ratio.}
    \label{fig:ablation-budget}
\end{figure}

To evaluate the Adaptive Budget Training strategy, we compared two settings: (1) Fixed Budget Training, where the model is trained exclusively with a maximum budget of $K=64$; and (2) Adaptive Budget Training, where $K$ is uniformly sampled from $\{8, 16, \dots, 64\}$. We assessed generalization capabilities using a series of unseen budget values $K \in \{10, 20, \dots, 60\}$, as shown in~\autoref{fig:ablation-budget}. The fixed budget model suffers from severe overfitting; it fails to retrieve historical information when the inference budget (e.g., $K=10 \text{-} 30$) is significantly smaller than the training budget, recovering only when $K$ approaches 64. In contrast, the adaptive model demonstrates strong robustness: it retains retrieval capabilities even at an extremely low budget ($K=10$) and performance improves consistently as the budget increases. This confirms that dynamic sampling enables the model to compress information adaptively based on available "bandwidth," rather than relying on a fixed number of positional encodings.

\subsection{Efficiency Analysis}
We further evaluate the computational efficiency of SONIC compared to the Full-context baseline. We conducted a performance test using a 30-turn long conversation sample. The results, summarized in~\autoref{tab:efficiency}, highlight the trade-offs and benefits of our approach.

\begin{table}[t]
\small
\centering
\caption{Efficiency Comparison between SONIC and Full-context Model on a 30-turn Sample. \textbf{Bold} indicates the best performance. $(\downarrow)$ denotes lower is better.}
\begin{tabularx}{\columnwidth}{l@{\extracolsep{\fill}}rrr}
\toprule
\textbf{Metric} & \textbf{Full} & \textbf{SONIC} & \textbf{$\Delta$ (\%)} \\ 
\midrule
Total Time (s) $\downarrow$     & 398.72 & \textbf{199.12} & $-50.1$ \\
Peak Memory (GB) $\downarrow$   & 10.95  & \textbf{3.58}   & $-67.3$ \\
Speed (tokens/s) $\uparrow$     & 27.65  & \textbf{30.36}  & $+9.8$  \\ 
\midrule
Nexus Gen. Time (s)             & ---    & 5.29            & ---     \\ 
\bottomrule
\end{tabularx}
\label{tab:efficiency}
\end{table}

Although the generation of Nexus tokens introduces a slight computational overhead, this cost is negligible compared to the overall gain in generation speed. SONIC achieves a higher generation speed and lower total inference time compared to the Full-context model. Furthermore, unlike standard KV cache compression strategies that may require re-computation based on the latest query, Nexus tokens are generated independently of the current query. This allows already generated Nexus tokens to be maintained throughout a multi-turn session, significantly reducing computational redundancy and memory occupation as the conversation progresses. Notably, SONIC also exhibits remarkable training efficiency; using an extremely compact synthetic dataset of only 4,949 samples, the entire training process converges within one hour on four NVIDIA A40 GPUs. This highlights SONIC as a lightweight, plug-and-play solution that achieves high performance with minimal data and compute requirements.

\section{Conclusion}
In this paper, we introduced SONIC, a novel framework designed to alleviate the memory bottleneck in multi-turn large language model interactions. By establishing Nexus tokens as learnable memory anchors, SONIC effectively compresses monotonically growing historical contexts into compact semantic representations. Coupled with the adaptive budget training strategy and multi-level distillation objectives, the framework enables flexible inference-time memory adaptation without requiring model retraining. Extensive experiments across diverse multi-turn benchmarks demonstrate that SONIC achieves performance comparable to full-context models while offering substantial compression ratios, significantly outperforming heuristic eviction baselines. Our work provides a robust and efficient paradigm for deploying long-context interactive systems.

\section*{Impact Statement}
This work aims to improve the efficiency of Large Language Models, enabling the deployment of long-context capabilities on resource-constrained hardware. By reducing the memory footprint of multi-turn dialogues, our method contributes to lowering the energy consumption of AI services and democratizing access to advanced conversational agents. We do not foresee any specific negative societal consequences that require highlighting here.

\nocite{langley00}

\bibliography{example_paper}
\bibliographystyle{icml2026}

\newpage
\appendix
\onecolumn

\section{Limitations}
Although SONIC demonstrates significant effectiveness in compressing KV caches for multi-turn dialogues, our evaluation is currently limited to dense Transformer models up to 4B parameters. The scalability of the Nexus mechanism on significantly larger models (e.g., 70B+) remains to be verified, as their attention patterns may exhibit higher complexity. Furthermore, we have not yet explored the applicability of SONIC to Mixture-of-Experts (MoE) architectures. Given the growing prevalence of sparse models in long-context reasoning, integrating Nexus-based compression with MoE represents a promising direction for future research.

\section{More Related Works}
\subsection{KV Cache Compression}
SnapKV~\cite{li2024snapkv} further refines this by selecting clustered attention patterns. Complementary to eviction, KV cache quantization compresses stored keys/values without removing tokens: KIVI~\cite{liu2024kivi} applies asymmetric 2-bit quantization, KVQuant~\cite{hooper2024kvquant} targets ultra-long contexts with quantization-aware strategies, and No Token Left Behind~\cite{yang2024no} uses importance-aware mixed precision to retain low-precision KV entries. While eviction-based methods are efficient, they fundamentally rely on local or statistical heuristics to evict information. This "hard" removal often severs long-range semantic dependencies that may not be statistically significant in the past but are critical for future reasoning. Our SONIC framework differs by adopting a \textit{learning-based} compression approach. Instead of eviction, we compress historical segments into learnable Nexus tokens, utilizing supervision from a full-context teacher to preserve semantic integrity beyond simple attention statistics.

\subsection{Memory Mechanisms in LLMs}
Enhancing LLMs with extended memory has been a longstanding goal. External memory systems, exemplified by RAG and MemGPT~\cite{packer2023memgpt}, manage context via retrieval indices or OS-like virtual memory calls. While powerful, they introduce system complexity and non-differentiable retrieval steps. Internal soft memory approaches, such as Transformer-XL~\cite{dai2019transformerxl} and Compressive Transformers~\cite{Rae2019CompressiveTF}, maintain recurrent segment-level states, while Recurrent Memory Transformer~\cite{bulatov2022recurrent} employs a fixed set of memory tokens updated across segments. Prompt compression methods offer another route by directly shortening inputs, including Gist Tokens~\cite{mu2023learning} and LLMLingua/LongLLMLingua~\cite{jiang2023llmlingua,jiang2024longllmlingua}. More recently, AutoCompressors~\cite{chevalier-etal-2023-adapting} proposed fine-tuning models to summarize contexts into summary vectors.

SONIC distinguishes itself from these methods through its structural optimization for \textit{multi-turn dialogue}. Unlike generic text compressors, SONIC leverages the inherent turn-based structure of conversation (User/Assistant) to perform segmented compression. By integrating dynamic budgeting and hierarchical visibility masks, SONIC functions as an internal, structured memory specifically tailored for the variable-length and interactive nature of multi-turn applications.

\section{Problem Formulation}
Consider a multi-turn conversation sequence $\mathcal{S} = \{T_1, T_2, \ldots, T_n\}$, where the $i$-th turn $T_i = (Q_i, R_i)$ consists of a user query $Q_i$ and an assistant response $R_i$. We denote the historical context preceding the current turn $n$ as $H_{<n} = \{I_{\text{sys}}, T_1, \ldots, T_{n-1}\}$, where $I_{\text{sys}}$ represents the system prompt.

In standard Transformer architectures, the model must maintain the KV cache for the entire history $H_{<n}$ to generate the response $R_n$. Consequently, the memory consumption of the KV cache scales linearly with the cumulative sequence length.

To alleviate this bottleneck, we introduce a structural compression mapping $\phi$. This function compresses the historical context states into a compact set of Nexus memory anchors. During inference, the effective context is composed of the system prompt $I_{\text{sys}}$, the Nexus state $\mathcal{N}$, and the current query $Q_n$. Our optimization objective is to maximize the conditional probability of the target response $R_n$:
\begin{equation}
    \begin{aligned}
       \max_{\theta}  & \quad  \log P_{\theta}(R_n \mid I_{\text{sys}}, N, Q_n), \\
        &\text{s.t.} \quad h \times K \le B,
    \end{aligned}
    \label{eq:objective}
\end{equation}
where $\theta$ represents the model parameters, $\mathcal{N}$ represents the set of Nexus tokens, $h$ denotes the number of historical segments, $K$ is the Nexus budget per segment, and $B$ is the maximum allowable memory budget.

Distinct from approaches with a fixed compression rate, we formulate the memory budget $B$ as a dynamic variable during inference. This is controlled by the number of Nexus tokens, $K$. Consequently, the model is required not only to optimize performance under a static budget but also to maintain robustness across varying compression levels defined by $K$. This formulation establishes the theoretical basis for the subsequent dynamic budget training strategy.

\section{Implementation Details}
\label{appendix:details}
All experiments were conducted on 4 NVIDIA A40 (40GB) GPUs. We fine-tuned the model on a specifically synthesized multi-turn conversation dataset designed to cover core capabilities, including status updating, constraint accumulation, coreference resolution, nested logic, and scattered information aggregation. To preserve the general capabilities of the original model, we only update Nexus-related parameters (Nexus Embeddings, Projection layers, and MLP modules).

During the training phase, the maximum Nexus token budget is set to 64. To facilitate dynamic budget adaptation, we randomly sample the Nexus token quantity $K$ from the discrete set $\{4, 8, 16, 32, 64\}$ at each iteration. The reconstruction loss weight is set to 1.0, and the logit distillation weight is set to 1.0 (with temperature $\tau=3.0$). The Nexus attention regularization weight is set to 0.5 (with threshold $\gamma=0.1$). Additionally, both the hidden-state alignment weight and the attention-guided distillation weight are set to 0.5, applied exclusively to the final layer of the model.

\section{Dataset Construction}
\label{appendix:data}
\subsection{Training Dataset}
To equip the model with robust long-context capabilities, we constructed a synthetic multi-turn dialogue dataset focusing on five core competencies: \textbf{Status Updating}, \textbf{Constraint Accumulation}, \textbf{Coreference Resolution}, \textbf{Nested Logic}, and \textbf{Scattered Information Aggregation}. The data generation pipeline consisted of multiple stages, where an advanced LLM (e.g. Qwen-Max) was prompted to generate complex conversation flows embedding these specific patterns. For instance, in the "Status Updating" category, the model simulates a scenario (e.g., a text-based game or inventory management) where the state changes dynamically across turns, requiring the model to track the latest valid state. We generated data for five stages. The final dataset ensures that the model learns to compress and retrieve information based on semantic structure rather than mere proximity.

\subsection{Benchmark Dataset}
\subsubsection{Dataset Statistics}
We summarize the key statistics of the four evaluation benchmarks in~\autoref{tab:dataset_stats}. These datasets vary significantly in scale, sequence length, and structural complexity, providing a comprehensive evaluation of the model's compression and retrieval capabilities.

\begin{table}[h]
    \centering
    \small
    \caption{Statistics of evaluation benchmarks. ``Avg. Tokens/Turn'' refers to the combined tokens of the user query and the assistant response. ``Avg. Hist. Tokens'' denotes the historical context length during the final response generation phase.}
    \begin{tabular}{lrrrr}
        \toprule
        Benchmark & Samples & Avg. Turns & Avg. Tokens/Turn & Avg. Hist. Tokens \\
        \midrule
        MTBench-101~\cite{bai2024mt} & 1,388 & 3.03 & 84.03 & 171.46 \\
        GSM8K-Variant~\cite{cobbe2021gsm8k} & 1,001 & 14.44 & 304.24 & 4,394.32 \\
        CoreRes & 978 & 7.00 & 313.31 & 2,193.85 \\
        SafeDialBench~\cite{cao2025safedialbenchfinegrainedsafetybenchmark} & 2037 & 4.92 & 678.18 & 2404.59 \\
        \bottomrule
    \end{tabular}
    \label{tab:dataset_stats}
\end{table}

The generation protocols differ across these benchmarks to reflect distinct real-world scenarios. \textbf{SafeDialBench} and \textbf{MTBench-101} are evaluated through iterative turn-by-turn generation, and they employ ground-truth responses to update the history after each turn, ensuring that errors in early turns do not derail the evaluation of subsequent tasks. \textbf{GSM8K-Variant} and \textbf{CoreRes} are designed as single-turn retrieval tasks where the model processes the entire historical dialogue at once to generate a single final answer, emphasizing the extraction of scattered information from extremely long multi-turn contexts.

\subsubsection{GSM8K-Variant Details}
We derived this benchmark from the test split of GSM8K~\cite{cobbe2021gsm8k}. The construction process involves two steps: \textbf{Decomposition} and \textbf{Scattering}. First, we utilized an LLM to decompose each math word problem into 2--4 independent atomic facts (conditions) and a final question. For example,``Tom has 2 apples and Jerry has twice as many" is split into ``1. Tom has 2 apples" and ``2. Jerry has twice as many as Tom". Second, we dispersed these atomic facts across a multi-turn conversation. To simulate a realistic and challenging retrieval environment, we interleaved significant amounts of topic-irrelevant "chitchat" or distraction turns between the condition segments. The model must retrieve all scattered conditions from the compressed history to answer the final question correctly.

\subsubsection{CoreRes Details}
Coreference Resolution (CoreRes) is a purely synthetic benchmark designed to test long-range entity tracking. We defined a set of diverse scenarios (e.g., "Cyberpunk Black Market", "Medieval Potion Shop") and target attributes (e.g., "Serial Number", "Magic Property"). For each sample, we generated a "seed" interaction:
\begin{itemize}
    \item \textbf{Injection Turn:} The user introduces an entity with a specific attribute value (e.g., ``The street price of this chip is 500 credits").
    \item \textbf{Distraction Turns:} A variable number of irrelevant dialogue turns are inserted to increase the context length and memory difficulty.
    \item \textbf{Retrieval Turn:} The user asks for the attribute value using an implicit reference or pronoun (e.g., "Wait, how much did that thing cost?"), strictly avoiding the entity's name to force the model to resolve the coreference through history.
\end{itemize}

\section{More Results}
\subsection{MTBench-101}
\begin{table*}[htbp]
\centering
\small
\caption{Detailed MTBench101 results under 80\% KV cache compression rate. \textbf{Bold} indicates the best performance among compression methods (excluding Full-context).}
\setlength{\tabcolsep}{4pt}
\renewcommand{\arraystretch}{1.1}
\begin{tabular}{ll|ccccccccccccc|c}
\hline
\textbf{Model} & \textbf{Method} & \textbf{GR} & \textbf{IC} & \textbf{AR} & \textbf{FR} & \textbf{MR} & \textbf{CC} & \textbf{TS} & \textbf{CR} & \textbf{SA} & \textbf{SI} & \textbf{CM} & \textbf{PI} & \textbf{SC} & \textbf{Avg} \\ \hline
\multirow{6}{*}{Qwen3-0.6B} & Full-context & 3.17 & 5.85 & 7.96 & 9.43 & 5.15 & 6.12 & 5.76 & 9.52 & 5.23 & 4.15 & 6.70 & 4.83 & 4.17 & 6.00 \\ \cline{2-16}
 & \myhl{\textbf{SONIC} (Ours)} & \myhl{2.90} & \myhl{4.29} & \myhl{4.76} & \myhl{6.09} & \myhl{3.26} & \myhl{4.19} & \myhl{3.51} & \myhl{5.95} & \myhl{3.22} & \myhl{3.04} & \myhl{3.76} & \myhl{3.14} & \myhl{2.94} & \myhl{\textbf{3.93}}  \\
 & StreamingLLM & 1.00 & 1.02 & 5.86 & 4.12 & 1.00 & 1.04 & 1.00 & 5.95 & 4.23 & 1.02 & 4.58 & 1.00 & 3.74 & 2.74 \\
 & ExpectedAttention & 1.00 & 1.00 & 6.82 & 8.11 & 1.00 & 1.06 & 1.01 & 8.79 & 6.12 & 1.03 & 5.09 & 1.00 & 5.06 & 3.62 \\
 & H2O & 1.00 & 1.00 & 6.92 & 8.46 & 1.00 & 1.04 & 1.02 & 8.69 & 5.66 & 1.02 & 5.00 & 1.00 & 5.82 & 3.66 \\
 & SnapKV & 1.00 & 1.01 & 5.39 & 6.38 & 1.00 & 1.03 & 1.00 & 7.54 & 5.51 & 1.01 & 4.70 & 1.00 & 5.12 & 3.21 \\ \hline
\multirow{6}{*}{Qwen3-1.7B} & Full-context & 5.59 & 7.48 & 9.64 & 9.88 & 6.99 & 8.42 & 7.88 & 9.89 & 8.48 & 6.12 & 9.03 & 7.37 & 8.68 & 8.11 \\ \cline{2-16}
 & \myhl{\textbf{SONIC} (Ours)} & \myhl{3.79} & \myhl{7.17} & \myhl{8.14} & \myhl{7.38} & \myhl{4.93} & \myhl{6.65} & \myhl{6.84} & \myhl{8.31} & \myhl{5.85} & \myhl{5.14} & \myhl{6.31} & \myhl{5.14} & \myhl{6.55} & \myhl{\textbf{6.32}} \\
 & StreamingLLM & 1.00 & 1.02 & 7.71 & 6.08 & 1.00 & 1.03 & 1.02 & 8.42 & 7.44 & 1.03 & 7.56 & 1.00 & 4.71 & 3.77 \\
 & ExpectedAttention & 1.00 & 1.01 & 9.13 & 9.36 & 1.01 & 1.04 & 1.00 & 9.73 & 9.01 & 1.03 & 8.90 & 1.00 & 8.39 & 4.74 \\
 & H2O & 1.03 & 1.01 & 9.28 & 9.62 & 1.00 & 1.05 & 1.01 & 9.79 & 8.85 & 1.02 & 8.60 & 1.01 & 8.29 & 4.74 \\
 & SnapKV & 1.00 & 1.03 & 7.52 & 7.46 & 1.02 & 1.05 & 1.01 & 9.51 & 7.29 & 1.01 & 7.66 & 1.00 & 8.10 & 4.21 \\ \hline
\multirow{6}{*}{Qwen3-4B} & Full-context & 8.10 & 8.53 & 9.95 & 9.97 & 8.37 & 9.50 & 8.93 & 9.88 & 9.07 & 7.15 & 9.66 & 8.48 & 9.82 & 9.03 \\ \cline{2-16}
 & \myhl{\textbf{SONIC} (Ours)} & \myhl{6.77} & \myhl{8.35} & \myhl{9.03} & \myhl{8.87} & \myhl{7.45} & \myhl{8.67} & \myhl{8.82} & \myhl{8.96} & \myhl{8.79} & \myhl{7.77} & \myhl{8.26} & \myhl{7.64} & \myhl{8.91} & \myhl{\textbf{8.33}} \\
 & StreamingLLM & 1.00 & 1.02 & 8.52 & 6.14 & 1.01 & 1.02 & 1.00 & 9.15 & 7.93 & 1.01 & 8.33 & 1.00 & 6.34 & 4.11 \\
 & ExpectedAttention & 1.00 & 1.01 & 9.89 & 9.74 & 1.00 & 1.05 & 1.00 & 9.90 & 9.63 & 1.03 & 9.56 & 1.00 & 9.45 & 5.02 \\
 & H2O & 1.00 & 1.00 & 9.88 & 9.78 & 1.02 & 1.04 & 1.00 & 9.85 & 9.48 & 1.03 & 9.60 & 1.00 & 9.32 & 5.00 \\
 & SnapKV & 1.00 & 1.02 & 8.22 & 8.20 & 1.03 & 1.04 & 1.01 & 9.81 & 7.48 & 1.01 & 8.09 & 1.01 & 7.08 & 4.31 \\ \hline
\end{tabular}
\label{tab:mtbench_0.8}
\end{table*}

\begin{table*}[htbp]
\centering
\small
\caption{Detailed MTBench101 results under 50\% KV cache compression rate. \textbf{Bold} indicates the best performance among compression methods (excluding Full-context).}
\setlength{\tabcolsep}{4pt}
\renewcommand{\arraystretch}{1.1}
\begin{tabular}{ll|ccccccccccccc|c}
\hline
\textbf{Model} & \textbf{Method} & \textbf{GR} & \textbf{IC} & \textbf{AR} & \textbf{FR} & \textbf{MR} & \textbf{CC} & \textbf{TS} & \textbf{CR} & \textbf{SA} & \textbf{SI} & \textbf{CM} & \textbf{PI} & \textbf{SC} & \textbf{Avg} \\ \hline
\multirow{6}{*}{Qwen3-0.6B} & Full-context & 3.17 & 5.85 & 7.96 & 9.43 & 5.15 & 6.12 & 5.76 & 9.52 & 5.23 & 4.15 & 6.70 & 4.83 & 4.17 & 6.00 \\ \cline{2-16}
 & \myhl{\textbf{SONIC}} & \myhl{3.14} & \myhl{5.57} & \myhl{7.22} & \myhl{7.69} & \myhl{4.24} & \myhl{5.51} & \myhl{5.53} & \myhl{8.49} & \myhl{4.66} & \myhl{4.17} & \myhl{6.29} & \myhl{4.97} & \myhl{5.47} & \myhl{\textbf{5.61}} \\
 & StreamingLLM & 1.00 & 1.01 & 6.65 & 8.36 & 1.01 & 1.03 & 1.00 & 8.15 & 5.73 & 1.03 & 5.34 & 1.00 & 5.22 & 3.58 \\
 & ExpectedAttention & 1.00 & 1.01 & 7.63 & 9.38 & 1.00 & 1.05 & 1.00 & 9.07 & 6.27 & 1.04 & 5.60 & 1.00 & 5.91 & 3.92 \\
 & H2O & 1.00 & 1.02 & 7.33 & 9.30 & 1.02 & 1.03 & 1.01 & 9.29 & 6.21 & 1.01 & 5.58 & 1.00 & 6.14 & 3.92 \\
 & SnapKV & 1.00 & 1.02 & 7.31 & 8.86 & 1.01 & 1.04 & 1.00 & 8.73 & 7.33 & 1.03 & 5.55 & 1.00 & 6.25 & 3.93 \\ \hline
\multirow{6}{*}{Qwen3-1.7B} & Full-context & 5.59 & 7.48 & 9.64 & 9.88 & 6.99 & 8.42 & 7.88 & 9.89 & 8.48 & 6.12 & 9.03 & 7.37 & 8.68 & 8.11 \\ \cline{2-16}
 & \myhl{\textbf{SONIC}} & \myhl{3.76} & \myhl{7.01} & \myhl{8.65} & \myhl{8.74} & \myhl{4.30} & \myhl{6.57} & \myhl{6.95} & \myhl{8.54} & \myhl{5.59} & \myhl{5.13} & \myhl{6.34} & \myhl{5.06} & \myhl{6.86} & \myhl{\textbf{6.42}} \\
 & StreamingLLM & 1.00 & 1.01 & 9.02 & 9.45 & 1.00 & 1.03 & 1.01 & 9.73 & 9.08 & 1.04 & 8.48 & 1.10 & 8.29 & 4.71 \\
 & ExpectedAttention & 1.01 & 1.03 & 9.54 & 9.74 & 1.00 & 1.05 & 1.00 & 9.88 & 9.11 & 1.01 & 9.01 & 1.00 & 8.49 & 4.84 \\
 & H2O & 1.00 & 1.01 & 9.41 & 9.58 & 1.04 & 1.05 & 1.00 & 9.91 & 9.16 & 1.01 & 9.23 & 1.02 & 8.96 & 4.88 \\
 & SnapKV & 1.01 & 1.02 & 9.24 & 9.70 & 1.01 & 1.03 & 1.00 & 9.79 & 8.95 & 1.01 & 8.68 & 1.00 & 8.47 & 4.76 \\ \hline
\multirow{6}{*}{Qwen3-4B} & Full-context & 8.10 & 8.53 & 9.95 & 9.97 & 8.37 & 9.50 & 8.93 & 9.88 & 9.07 & 7.15 & 9.66 & 8.48 & 9.82 & 9.03 \\ \cline{2-16}
 & \myhl{\textbf{SONIC}} & \myhl{6.86} & \myhl{8.44} & \myhl{9.12} & \myhl{8.96} & \myhl{7.55} & \myhl{8.77} & \myhl{8.92} & \myhl{9.06} & \myhl{8.89} & \myhl{7.87} & \myhl{8.36} & \myhl{7.74} & \myhl{9.01} & \myhl{\textbf{8.43}}  \\
 & StreamingLLM & 1.00 & 1.00 & 9.61 & 9.84 & 1.00 & 1.02 & 1.01 & 9.85 & 9.47 & 1.03 & 9.15 & 1.00 & 9.70 & 4.98 \\
 & ExpectedAttention & 1.00 & 1.01 & 9.88 & 9.91 & 1.00 & 1.05 & 1.00 & 9.93 & 9.62 & 1.04 & 9.61 & 1.00 & 9.69 & 5.06 \\
 & H2O & 1.00 & 1.01 & 9.84 & 9.78 & 1.00 & 1.05 & 1.00 & 9.95 & 9.41 & 1.03 & 9.64 & 1.00 & 9.77 & 5.04 \\
 & SnapKV & 1.00 & 1.01 & 9.71 & 9.84 & 1.00 & 1.05 & 1.02 & 9.85 & 9.59 & 1.02 & 9.48 & 1.01 & 9.42 & 5.00 \\ \hline
\end{tabular}
\label{tab:mtbench_0.5}
\end{table*}

\end{document}